\documentclass{article}

\usepackage{microtype}
\usepackage{graphicx}
\usepackage{subcaption}
\usepackage{booktabs}
\usepackage{hyperref}

\usepackage[accepted]{icml2026}

\usepackage{amsmath}
\usepackage{amssymb}
\usepackage{mathtools}
\usepackage{amsthm}
\usepackage{float}

\usepackage[capitalize,noabbrev]{cleveref}

\theoremstyle{plain}

\theoremstyle{definition}

\theoremstyle{remark}

\usepackage[textsize=tiny]{todonotes}

\icmltitlerunning{Quantifying Empirical Compute-Supervision Tradeoffs in RLVR}

\begin{document}

\twocolumn[
  \icmltitle{Quantifying Empirical Compute-Supervision Tradeoffs in RLVR}

  \icmlsetsymbol{equal}{*}

  \begin{icmlauthorlist}
    \icmlauthor{Ryo Mitsuhashi}{equal,princeton}
    \icmlauthor{Patrick Chen}{equal,princeton}
    \icmlauthor{Isabelle Tseng}{equal,princeton}
    \icmlauthor{Jasin Cekinmez}{equal,princeton}
    \icmlauthor{Addison J. Wu}{equal,princeton}
  \end{icmlauthorlist}

  \icmlaffiliation{princeton}{Princeton University}

  \icmlcorrespondingauthor{Ryo Mitsuhashi}{rm4411@princeton.edu}
  \icmlcorrespondingauthor{Patrick Chen}{patrickchen@princeton.edu}
  \icmlcorrespondingauthor{Isabelle Tseng}{isabelletseng@princeton.edu}
\icmlcorrespondingauthor{Jasin Cekinmez}{jasincekinmez@princeton.edu}
\icmlcorrespondingauthor{Addison J. Wu}{addisonwu@princeton.edu}

  \icmlkeywords{Machine Learning, ICML}

  \vskip 0.3in
]

\printAffiliationsAndNotice{\icmlEqualContribution}

\begin{abstract}
Reinforcement learning with verifiable rewards (RLVR) has become a standard paradigm for post-training language models, but in practice, verifiers are rarely perfect. Recent theoretical work predicts that verifier noise affects the rate of learning but not its final outcome, implying that sufficient compute should close any gap induced by imperfect supervision. We test this prediction empirically by post-training Qwen2.5 (0.5B, 1.5B) with GRPO on GSM8K while injecting controlled false-positive and false-negative noise into the binary correctness signal, and varying rollouts per prompt as a compute axis. In practice, the gap in validation accuracy persists under substantial compute scaling, with returns to compute that are sharply diminishing. We further find a structural asymmetry where false negatives monotonically degrade performance more quickly than false positives. These findings suggest verifier quality and training compute are not interchangeable, and that reducing false negatives is a more effective lever than scaling compute alone.\end{abstract}

\section{Introduction}

Reinforcement learning with verifiable rewards (RLVR) has recently emerged as a scalable paradigm for training language models without relying on costly human supervision \cite{lambert2024tulu3, guo2025deepseek}. Reward signals are produced by automatic evaluators such as symbolic checkers or unit tests, enabling large-scale optimization driven entirely by programmatic feedback \citep{le2022coderl, liu2023rltf}. However, in realistic settings, these verifiers are rarely perfect: they may produce false positives or false negatives, fail to capture all aspects of correctness, or provide delayed and sparse signals \cite{cai2025noisy,plesner2026imperfect}. Despite the growing reliance on such imperfect supervision, it remains unclear whether additional compute can offset errors in the verifier, or whether such errors ultimately cap or even degrade true performance.

In this work, we study the compute–supervision tradeoff in RLVR by jointly scaling post-training compute and verifier quality. Our central question is whether additional compute can compensate for degraded supervision, and whether this compensation is complete, partial, or fundamentally limited. Recent theoretical work predicts that verifier noise affects only the rate of learning, not its final outcome \citep{rad2026noisy}. To investigate the degree to which this finding transfers in practice, we treat verifier reliability as a controllable scaling variable alongside compute. We train language models on tasks with exact ground-truth evaluation, using a simple baseline setting of mathematical reasoning, while systematically introducing controlled imperfections into the training verifier and varying the total amount of compute that can be used. 

Our primary goal is to recover compute–verifier tradeoff curves that quantify how much additional compute is required to reach a fixed level of true performance under degraded supervision. By making this tradeoff explicit, we aim to clarify when RLVR can rely on imperfect automated feedback. Empirically, we find that performance improves only sub-linearly with additional compute and cannot close the gap induced by noisy verifiers, shedding light on practical limitations of previous theoretical claims stating that compute alone is sufficient.

\section{Related Work}

\subsection{Scaling Laws}
\begin{figure*}[t!]
    \centering
    \includegraphics[width=0.9\textwidth]{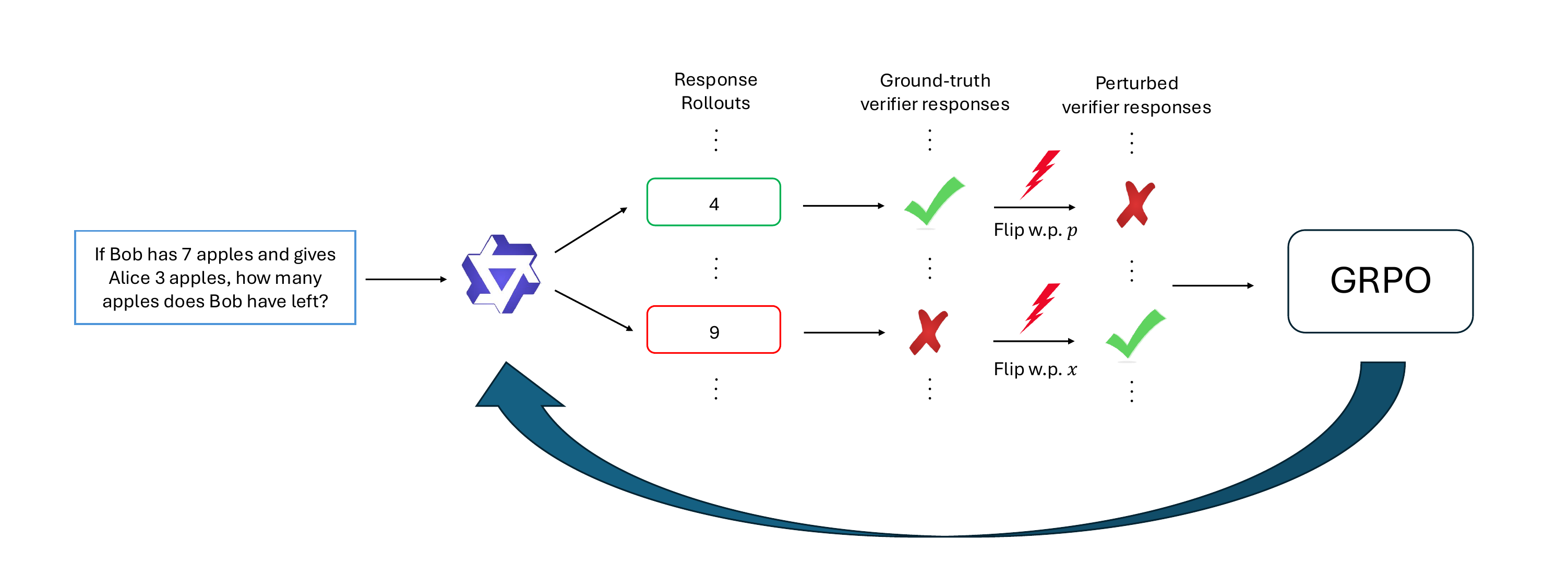}
    \caption{Noisy-reward GRPO pipeline. Given a prompt, the policy model samples a group of response rollouts that are scored by a ground-truth verifier. We then perturb the verifier signal by independently flipping correct rewards to incorrect with probability pp
p and incorrect rewards to correct with probability xx
x, yielding an asymmetrically noisy reward that is passed to GRPO for policy updates.}
    \label{fig:method}
\end{figure*}
Scaling laws are a central component of machine learning, providing a framework for understanding how model performance varies with compute, data, and model size. \citet{kaplan2020scaling} show that language model performance follow power law relationships with these factors. However, \citet{hoffmann2022empirical} demonstrate that many models are undertrained relative to their available data, highlighting the importance of properly balancing these variables. Given the importance of allocating compute and data effectively and the lack of strong understanding of scaling laws in RL settings, developing scaling laws under RLVR becomes paramount.

\subsection{Reinforcement Learning with Verifiable Rewards}
Verifiable rewards are signals with objective, checkable outputs that enable more stable reinforcement learning. However, many settings rely on noisier rewards for example RLHF \citep{yan2024reward}. \citet{rad2026noisy} characterize learning dynamics under noisy verification, showing that positive signal leads to improvement and proving that noise affects the \emph{rate} but not the asymptotic \emph{outcome} of learning. We complement this by quantifying how much compute is required to realize this equivalence in practice, and find that the gap persists well beyond realistic training budgets. Other work studies correcting noisy rewards during training \citep{cai2025noisy} or the limits of inference-time scaling under imperfect verifiers \citep{stroebl2026the}, but does not address training-time tradeoffs. In contrast, we study \textbf{training-time compute scaling under imperfect supervision}, directly testing whether increased optimization can offset degraded verifier quality or instead leads to persistent performance gaps.

\subsection{Reward Overoptimization and Hacking}
An adjacent line of work examines flawed signals in reward functions and their downstream consequences of optimizing to these signals. \citet{gao2023scaling} establish scaling laws for reward model overoptimization, showing that KL divergence from the initial policy predicts the gap between proxy and true reward as the policy exploits reward model weaknesses, a phenomenon formalized as reward hacking \citep{skalse2022defining, pan2022effects}. Recent work documents this in RLVR post-training for frontier LLMs. \citet{baker2025monitoring} show that optimization pressure against a CoT monitor produces obfuscated reward hacking rather than eliminating it, and \citet{metr2025rewardhacking} report that more capable models hack more aggressively. Our research question and setting differs in that we model the verifier as \emph{stochastically} noisy rather than systematically exploitable. Concretely, we ask whether additional training-time compute \emph{compensates} for degraded verifier quality rather than whether optimization induces drift against a fixed flawed objective.

\section{Methodology}

\subsection*{Training setup}
\begin{figure*}[t!]
    \centering
    \includegraphics[trim={0 0 0 2cm},clip,width=\linewidth]{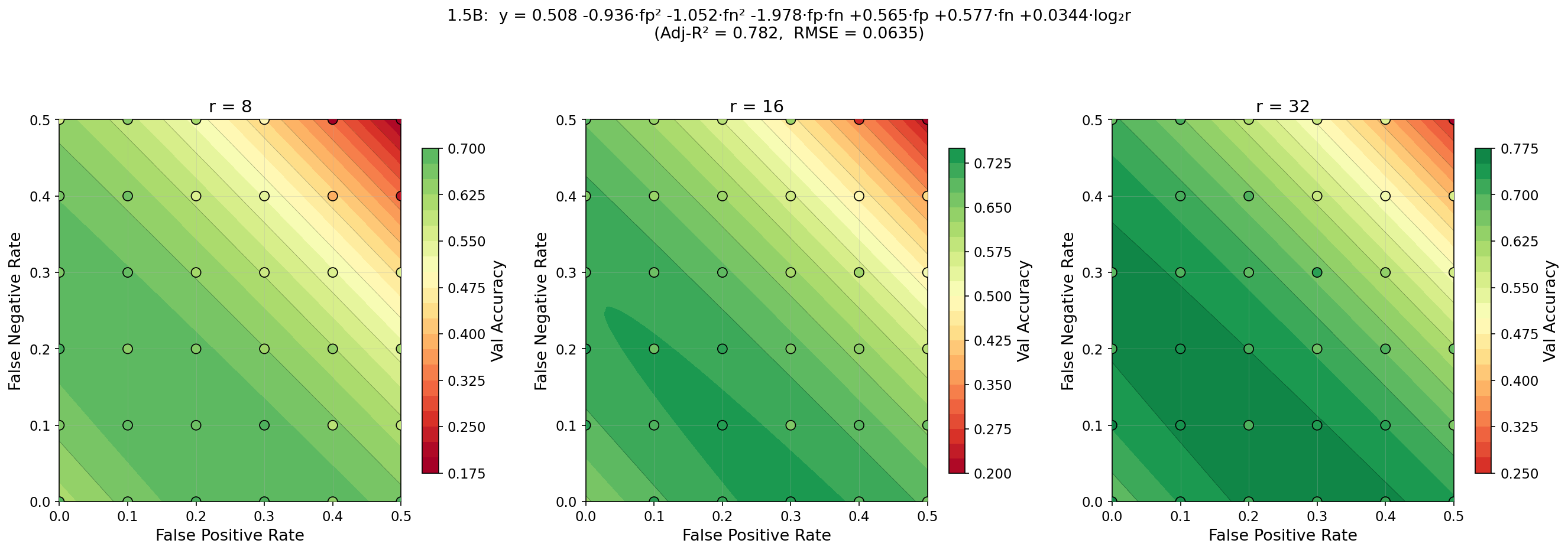}
    \caption{Validation accuracies for Qwen2.5 1.5B after GRPO post-training with verifiers of varying degradation, for rollout counts of 8, 16, 32, respectively, in the three sub-plots. Higher validation accuracies are obtained for lower false positive and negative rates, and this is also increased with additional rollouts. All heatmaps in this paper use the same universal color-accuracy scale. See Section \ref{subsec:scaling1.5} for the specific scaling trends.}
    \label{fig:1.5bresults}
\end{figure*}
\begin{figure*}[hbt!]
    \centering
    \includegraphics[trim={0 0 0 2cm},clip,width=\linewidth]{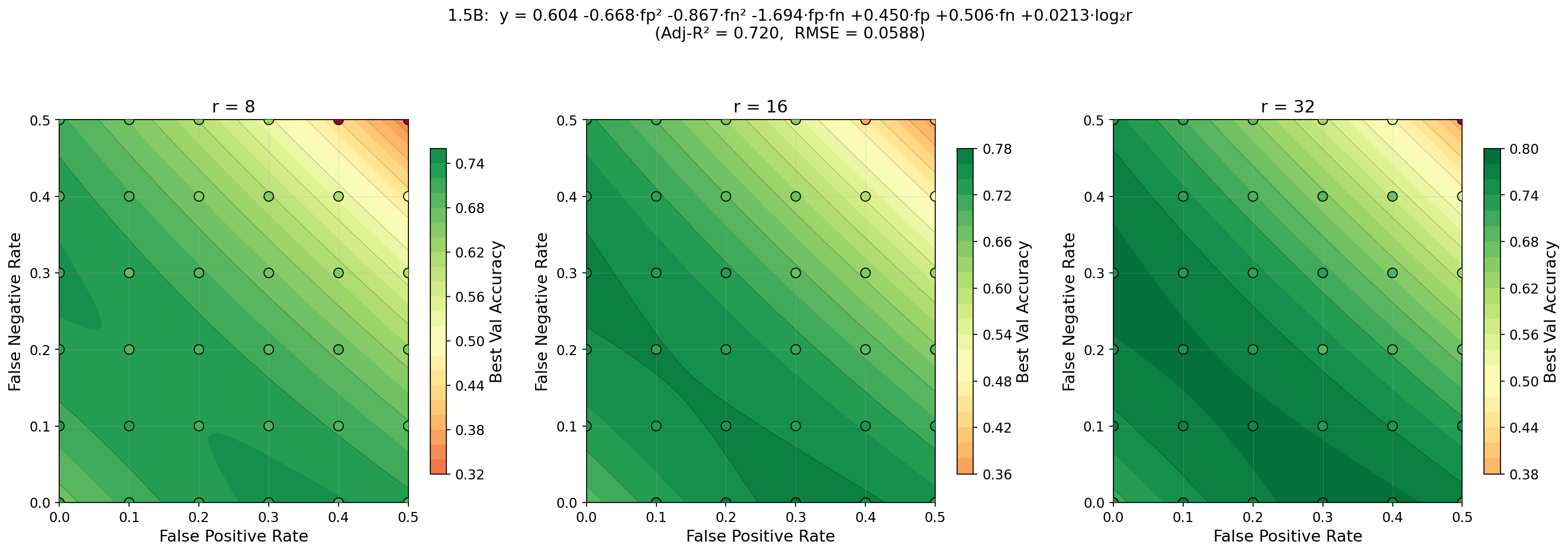}
    \caption{Scaling laws on 1.5B with best validation accuracy}
\end{figure*}
We post-train Qwen2.5 0.5B and 1.5B~\citep{qwen2025qwen25technicalreport} with GRPO~\citep{grpo} on GSM8K~\citep{gsm8k}, using the NeMo-RL codebase~\citep{nemo-rl}\footnote{Our code is available here: \url{https://github.com/rm-3284/RLVR_noisy_rewards.git}} modified to support a configurable noisy verifier. We chose GSM8K because its binary correctness signal provides a clean base reward into which we can inject controlled noise. All runs use AdamW with learning rate $5\times 10^{-6}$, weight decay $0.01$, betas $(0.9, 0.999)$, and gradient clipping at $1.0$. A linear warmup over $50$ steps (from factor $0.1$) is followed by a constant schedule. KL regularization against the reference policy uses a coefficient of $0.01$ with the k3 estimator, and PPO-style importance-ratio clipping at $0.2$. Generation is handled by vLLM~\citep{vllm} at temperature $1.0$.

\subsection*{Noisy verifier model}
We model an imperfect verifier as a stochastic perturbation of the ground-truth binary reward. Given the true correctness label $y^\star \in \{0,1\}$ produced by the GSM8K checker, the training reward is flipped independently with probability $p$ when $y^\star = 1$ (false negative) and with probability $x$ when $y^\star = 0$ (false positive). The effective reward is
\[
r = 
\begin{cases}
1 - y^\star & \text{w.p. } p \text{ if } y^\star = 1, \\
1 - y^\star & \text{w.p. } x \text{ if } y^\star = 0, \\
y^\star & \text{otherwise.}
\end{cases}
\]
as shown in Figure \ref{fig:method}. This parameterization lets us independently vary the false-negative rate $p$ and false-positive rate $x$, covering the full space of symmetric and asymmetric noise regimes. We sweep $p, x \in \{0, 0.1, 0.2, 0.3, 0.4, 0.5\}$, where $(p, x) = (0, 0)$ is the perfect verifier and $(0.5, 0.5)$ corresponds to completely random noise. Noise is applied only during training. All hold-out evaluations use the exact GSM8K verifier, so reported accuracies measure true capability rather than noisy reward.

\subsection*{Compute axis}
We treat the number of GRPO rollouts per prompt $r$ as our compute-scaling variable, sweeping $r \in \{8, 16, 32\}$. Preliminary experiments varying both rollouts and batch size showed that batch size had a negligible effect on the training curve while rollout count materially affected training stability and final performance, so we fix batch size at 32 and vary only $r$. This is consistent with the view that rollout count directly controls the effective number of gradient updates per unit of training data in GRPO, making it the natural compute axis for this setting.
\begin{figure}[t!]
    \centering
    \includegraphics[trim={0 0 0 1.2cm},clip,width=\columnwidth]{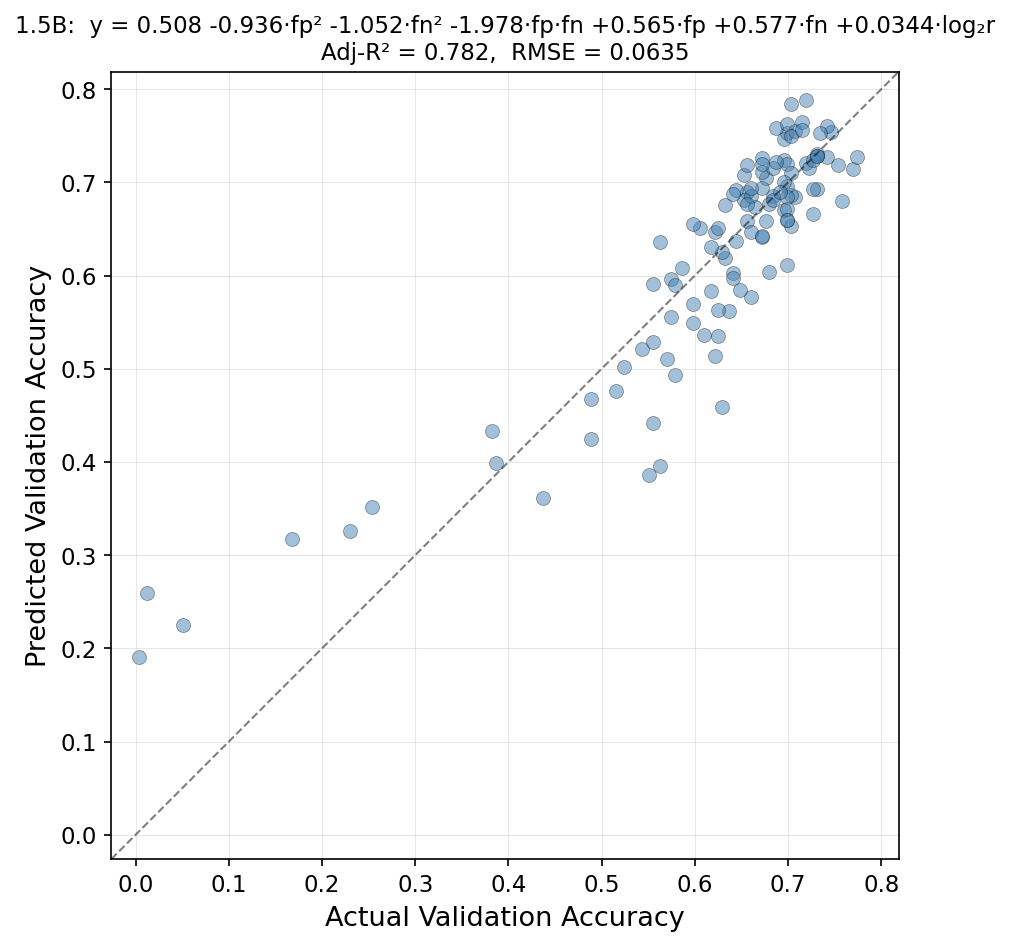}
    \caption{Predicted vs. actual final validation accuracy on 1.5B model. Adjusted $R^2 = 0.782$.}
    \label{fig:1.5bfinalfit}
\end{figure}
\subsection*{Training and evaluation protocol}
For each $(p, x, r)$ configuration, we run one seed for one epoch. We chose one epoch because preliminary runs showed that the training reward curve plateaus well before the end of an epoch across all noise levels, so additional epochs do not reflect further learning but only additional exposure to the same data under the same noise realization. We report both \emph{final-step} validation accuracy (main text) and \emph{best-step} validation accuracy across training (appendix) as a robustness check against late-training instability. For both the 1.5B model and 0.5B model, we ran $N = 108$ configurations across the full $(p, x, r)$ grid.

\section{Results}
\subsection*{Model details}
We model trained-model accuracy as a second-order polynomial in the verifier's false-positive rate $x$ and false-negative rate $p$, plus a log-linear term in rollouts $r$ to get validation accuracy $$y\approx a\cdot x^2 + b\cdot x\cdot p + c\cdot p^2 + d\cdot x + e\cdot p + f\log{r} + g$$
The polynomial part is a second-order Taylor expansion of an arbitrary smooth dependence on the two error rates around the perfect verifier point $(0, 0)$. Our $r$-values are geometrically spaced $r\in\{8, 16, 32\}$, so $\log{r}$ treats them as evenly spaced in the regressor; we do not claim to resolve the functional form in $r$ from only three values. We chose $\log{r}$ because it aligns with the standard practice of studying RL training effects on a multiplicative compute scale.

\subsection*{Scaling observations on 1.5B model}
\label{subsec:scaling1.5}
To provide better generalization of our results, we report our findings on the final validation accuracy for each configuration. Across our configurations on the 1.5B model, we have $N=108$ observations and get the curve \begin{multline*}
    y=- 0.936x^2 - 1.052p^2 - 1.978xp \\
    + 0.565x + 0.577p + 0.0344\log_2r + 0.508 
\end{multline*}
with adjusted $R^2 = 0.782$. See \Cref{fig:1.5bfinalfit} for the detailed fit. As visible from \Cref{fig:1.5bresults}, validation accuracy decreases as false positive and false negative rates increase, with strongest degradation when both are high. Increasing compute $(r=8\rightarrow 32)$ improves performance across all noise levels but cannot eliminate noise-induced gaps. False negatives additionally impact validation accuracy more than false positives.

\noindent Maximizing the fitted surface over the sampled region $[0, 0.5]^2$ yields an optimum at $(p, x)\approx (0, 0.3)$, with predicted accuracy roughly 0.07 higher than a perfect verifier $(0, 0)$. The optimum sits on the $p=0$ boundary, indicating that false negatives monotonically hurt training while a moderate rate of false positives is beneficial. This asymmetry is consistent with prior analyses in which false negatives deprive the policy of informative gradients, whereas false positives act as a form of reward perturbation that can regularize training~\citep{plesner2026imperfect}.

\noindent To ensure robustness, we also report our findings using the best validation accuracy in each configuration across all time steps. We observed a similar trend, including with the 0.5B model, and full results can be found in Appendix \ref{extended}.

\section{Conclusion}
We provide a compute-constrained perspective on RL post-training, showing that performance improves minimally with additional compute, leading to diminishing returns. As a result, increased compute cannot eliminate the performance degradation caused by noisy verifiers, and the gap persists under substantial scaling. These findings quantify the limits of compute as a substitute for verifier quality and suggest a more practical takeaway, improving the verifier particularly by reducing false negatives is a far more effective route for performance gains than scaling compute alone. 

Due to compute constraints, our post-training experiments are limited to relatively small Qwen models, however the observed trends are consistent with our underlying hypothesis and are likely to extend to larger and more diverse model families. Our experiments are conducted on GSM8K~\citep{gsm8k}, which consists of relatively simple arithmetic reasoning tasks. However, the fact that gaps still persist in this simpler setting likely suggest that such trends still hold in more complex settings where imperfect verifiers are more common. While we quantify the impact of noisy supervision, particularly false negatives as a key limitation, we do not propose mitigation strategies in this work, and developing methods to address these issues remains an important direction for future work. Lastly, we run one seed per configuration due to compute constraints. The overall trends are robust across the sweep, although individual configurations, especially the boundary optimum near $(p, x) \approx (0, 0.3)$, should be interpreted carefully.

\bibliography{example_paper}
\bibliographystyle{icml2026}

\newpage
\appendix
\setcounter{figure}{0}
\setcounter{table}{0}
\renewcommand{\thefigure}{A\arabic{figure}}
\renewcommand{\thetable}{A\arabic{table}}
\onecolumn

\section{Extended Results}\label{extended}

\begin{table}[hbt!]
    \centering
    \begin{tabular}{c|c|c|c}
        Model size & Final/Best & Fit equation & Adj.-$R^2$ \\
    \hline
        1.5B & Final & $- 0.936x^2 - 1.052p^2 - 1.978xp + 0.565x + 0.577p + 0.0344\log_2r + 0.508 $ & 0.782 \\
        1.5B & Best & $- 0.668x^2 - 0.867p^2 - 1.694xp + 0.450x + 0.506p +     0.0213\log_2{r} + 0.604$ & 0.720 \\
        0.5B & Final & $- 0.389x^2 - 0.617p^2 - 0.926xp - 0.088x + 0.114p + 0.0621\log_2r + 0.216 $ & 0.850 \\
        0.5B & Best & $- 0.386x^2 - 0.601p^2 - 0.990xp + 0.077x + 0.204p + 0.0317\log_2r + 0.366 $ & 0.862
    \end{tabular}
    \caption{Curves for each configuration}
    \label{tab:fits}
\end{table}
We show the full results for 1.5B and 0.5B in \Cref{tab:fits}.

\section{Supplementary Figures}

\begin{figure}[hbt!]
    \centering
    \includegraphics[trim={0 0 0 2cm},clip,width=0.7\linewidth]{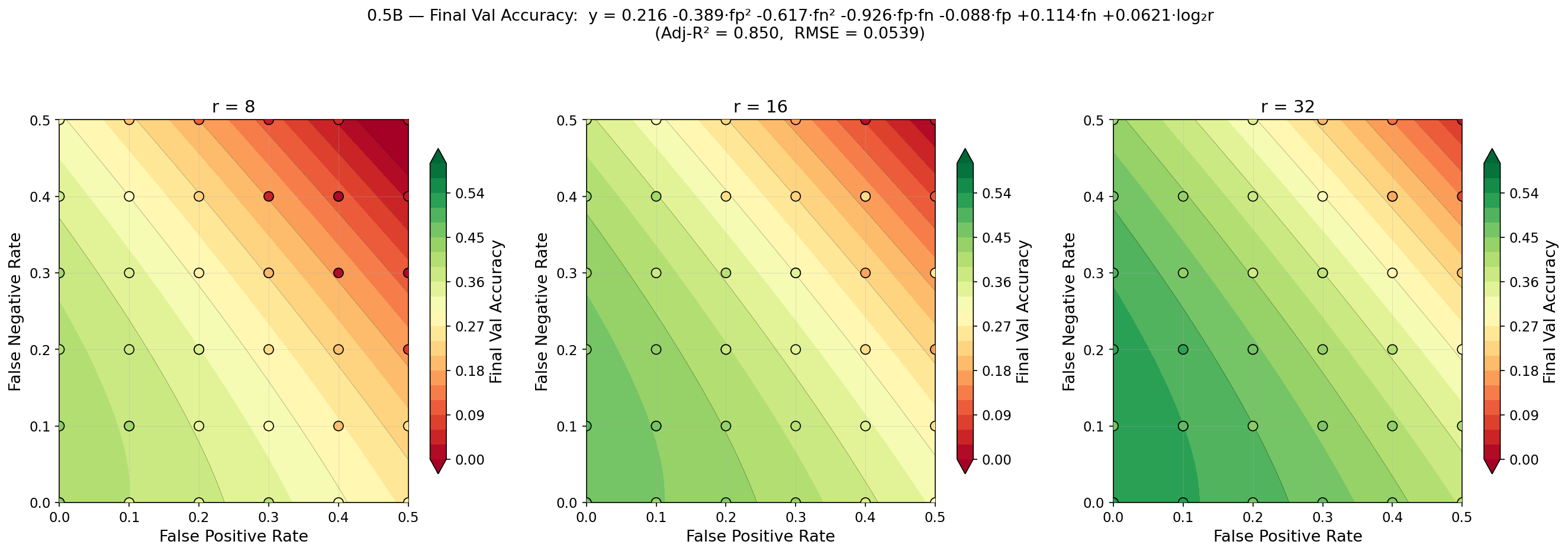}
    \caption{Scaling laws on 0.5B with final validation accuracy}
\end{figure}

\begin{figure}[hbt!]
    \centering
    \includegraphics[trim={0 0 0 1.2cm},clip,width=0.5\linewidth]{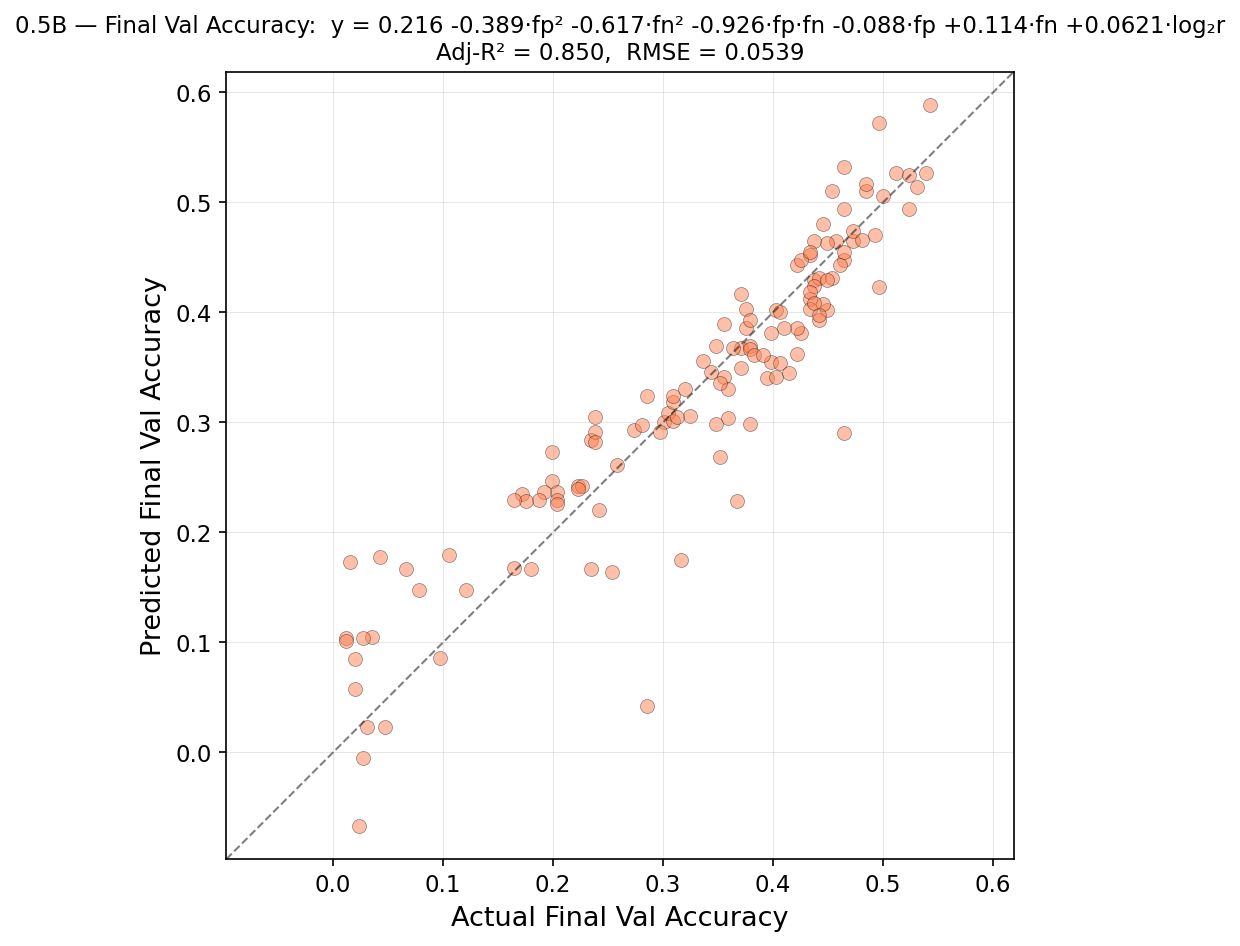}
    \caption{Predicted vs. Actual final validation accuracy on 0.5B model}
\end{figure}

\begin{figure}[hbt!]
    \centering
    \includegraphics[trim={0 0 0 2cm},clip,width=0.7\linewidth]{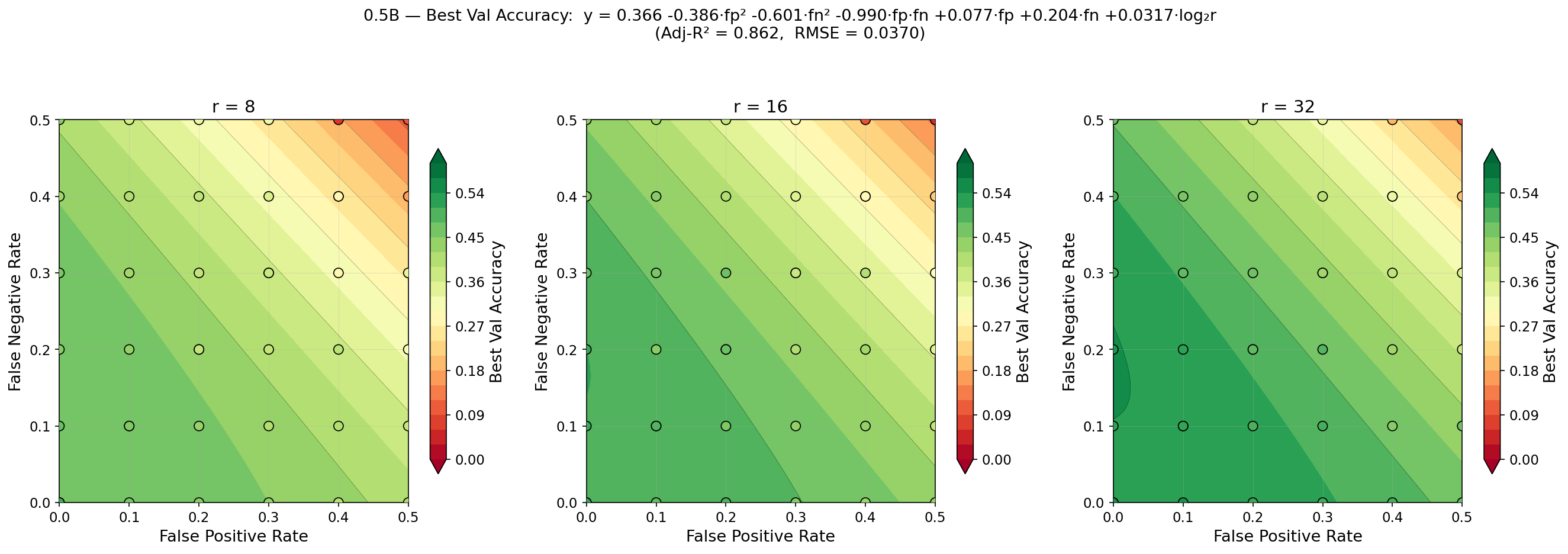}
    \caption{Scaling laws on 0.5B with best validation accuracy}
\end{figure}

\begin{figure}[hbt!]
    \centering
    \includegraphics[trim={0 0 0 1.2cm},clip,width=0.5\linewidth]{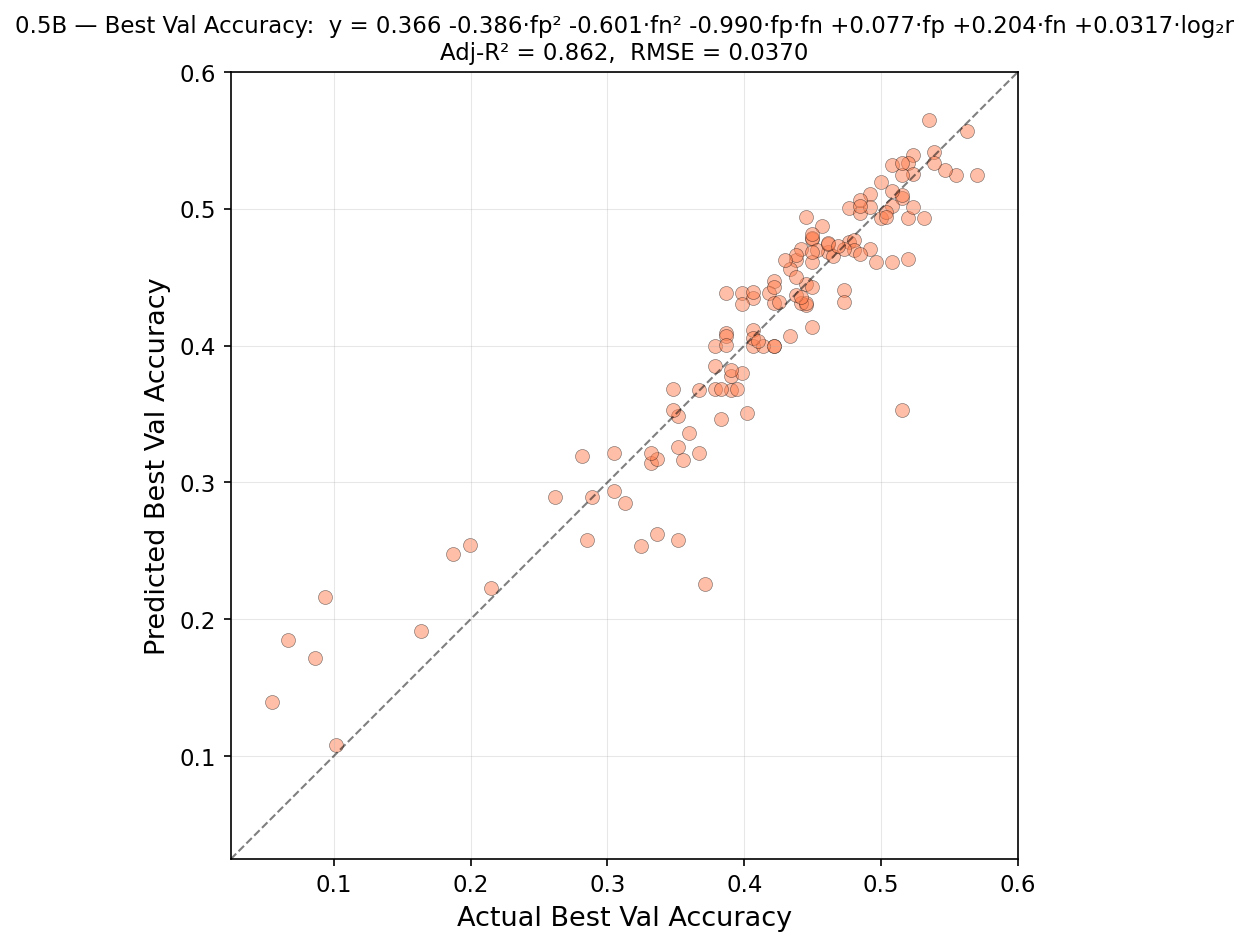}
    \caption{Predicted vs. Actual best validation accuracy on 0.5B model}
\end{figure}

\section{Analysis on Symmetric Error Cases}

On cases with symmetric error rates ($p = x$), we expand the number of rollouts to $r \in \{ 4, 8, 16, 32, 64 \}$.

\begin{figure}[h]
    \centering
    \includegraphics[width=0.9\linewidth]{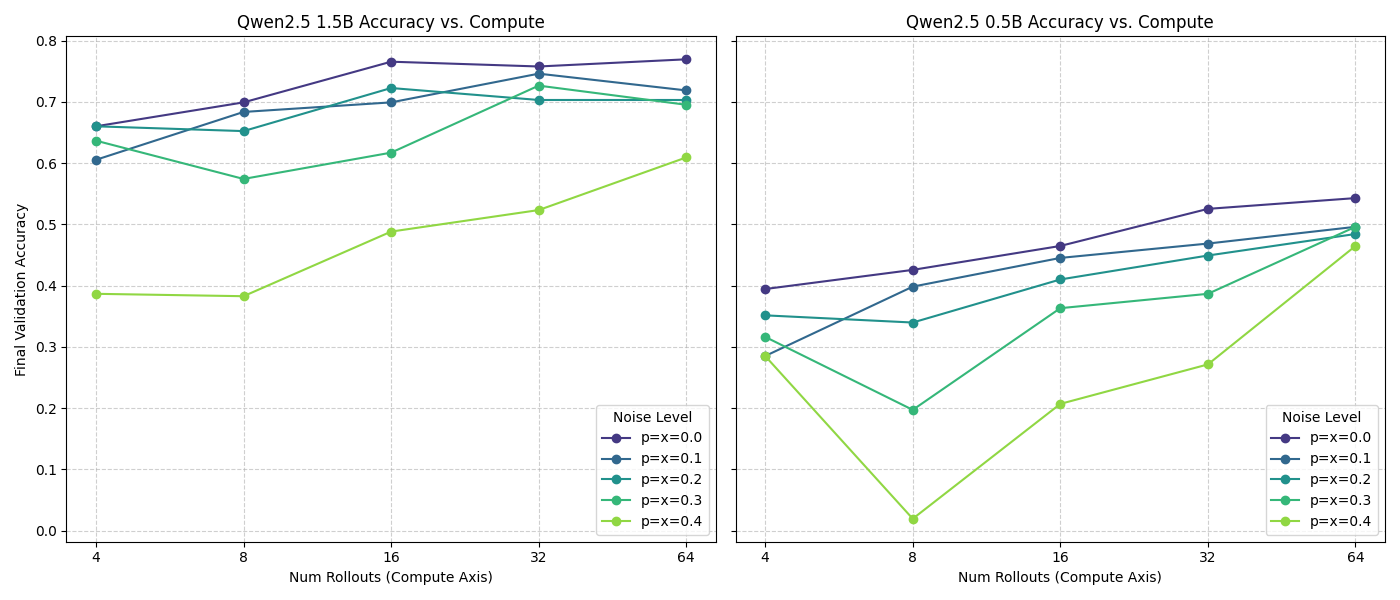}
    \caption{The relationship between the final accuracy and number of rollouts with symmetric noise. Qwen2.5 1.5B on the left and Qwen2.5 0.5B on the right.}
    \label{fig:rollouts}
\end{figure}

Figure~\ref{fig:rollouts} shows the final accuracy with various error rates and rollouts. In most cases, the final accuracy increases as you increase the number of rollouts. However, we could see some plateaus with the bigger model when the number of rollouts is large, and the error rate is low, suggesting that the model has reached its capacity limit. This also shows the shortcomings of our polynomial approximation, as it does not take into account the model size and treats every rollout increase (4 to 8 vs 32 to 64) equally. It is notable that in many cases, the increase in compute eventually compensates for the error rate, as $(p, r) = (0.3, 64)$ beats $(0.0, 4)$ in Qwen 1.5B and $(0.4, 64)$ beats $(0.0, 4)$ in Qwen 0.5B.

\begin{figure*}[t!]
    \centering
    \begin{subfigure}{0.49\linewidth}
        \centering
        \includegraphics[width=\linewidth]{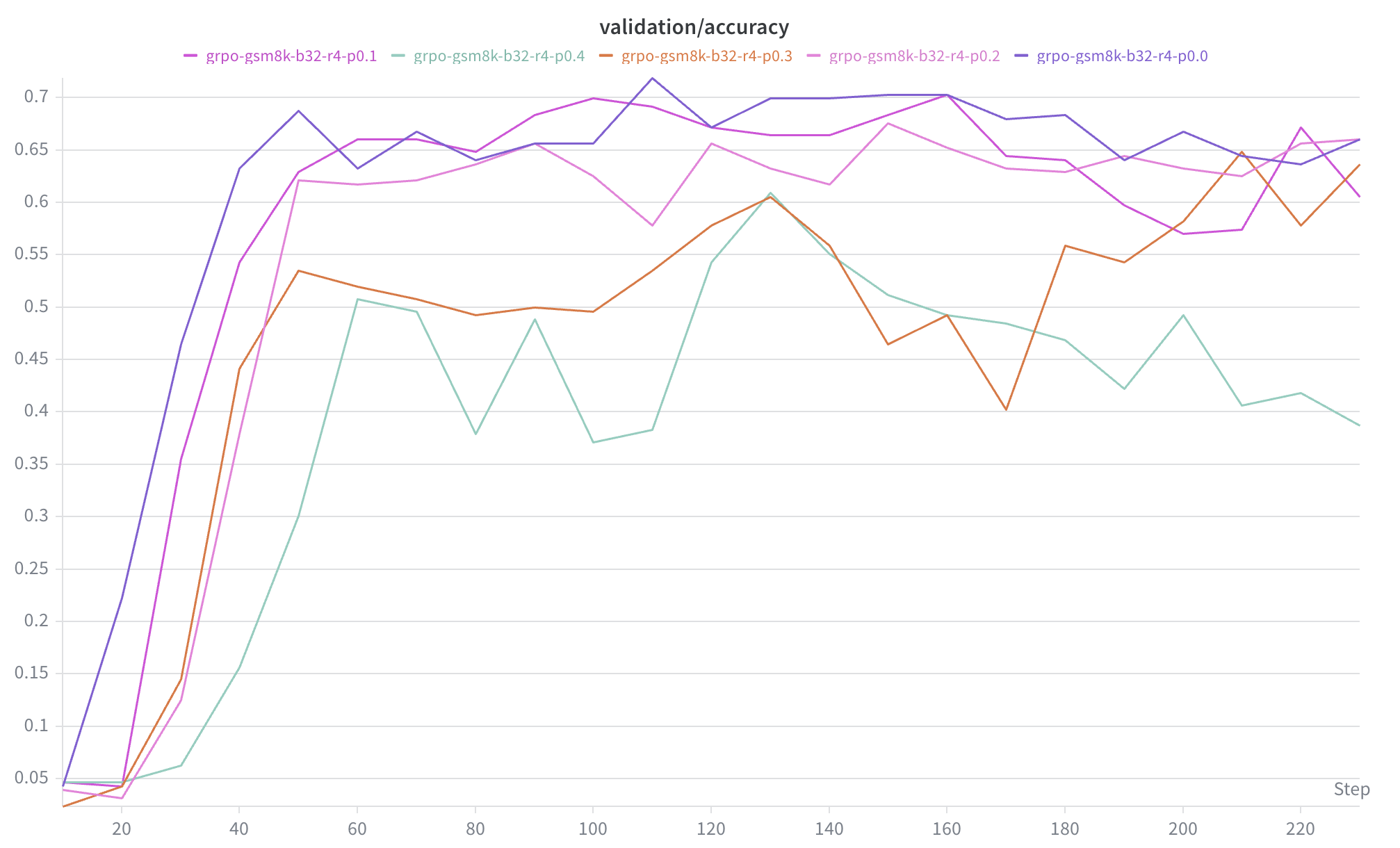}
        \caption{The learning curve of GRPO with rollout 4.}
        \label{fig:0.rollout4}
    \end{subfigure}
    \hfill
    \begin{subfigure}{0.49\linewidth}
        \centering
        \includegraphics[width=\linewidth]{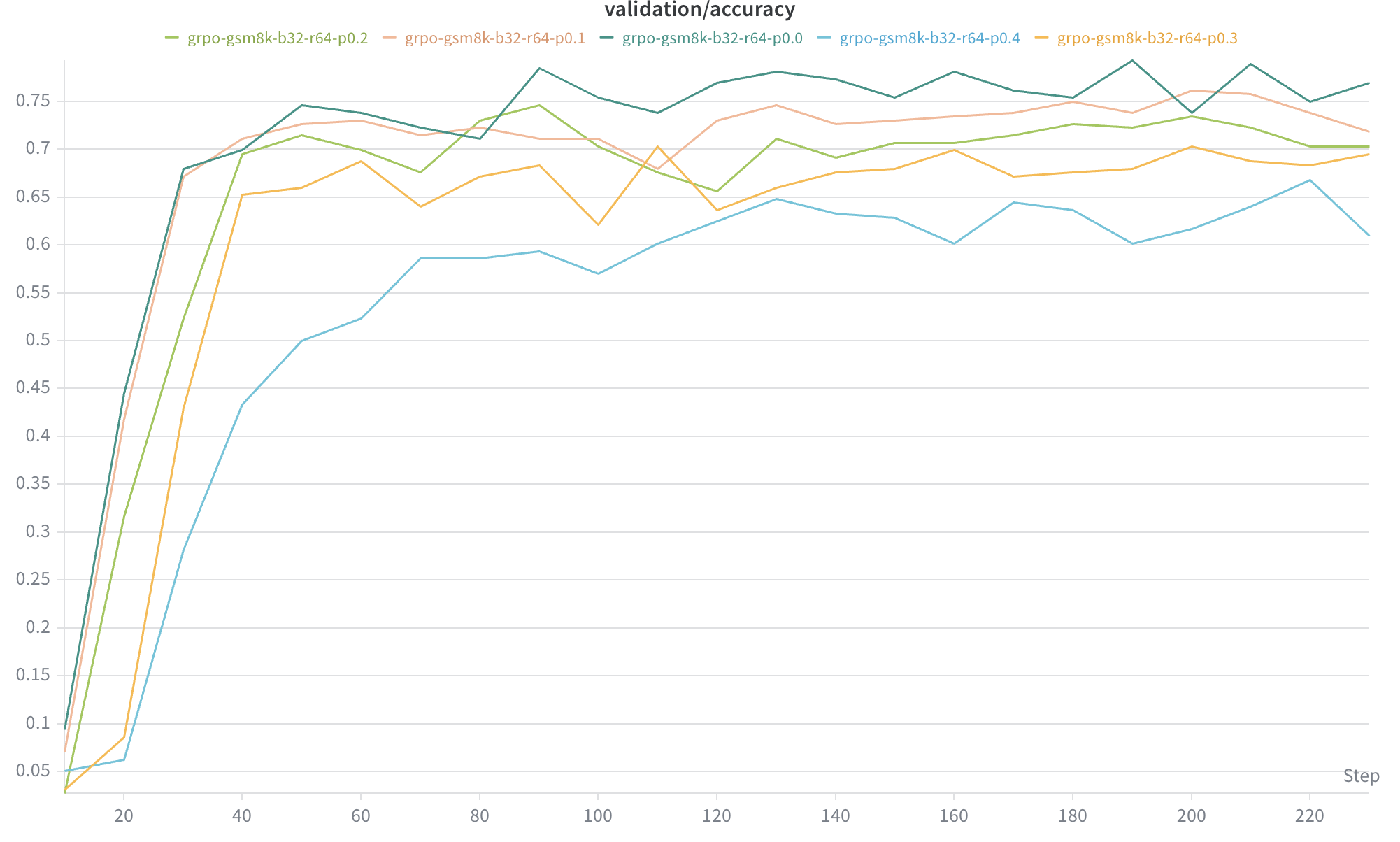}
        \caption{The learning curve of GRPO with rollout 64.}
        \label{fig:rollout64}
    \end{subfigure}
    \caption{The changes in validation accuracy throughout the training of Qwen2.5-1.5B with symmetric noise.}
    \label{fig:rollout_learning_curve}
\end{figure*}

Figure~\ref{fig:rollout_learning_curve} shows the accuracy curves of the GRPO with extreme rollouts, 4 and 64. Increasing the error rates not only affects the final accuracy but also the slope of the accuracy curve. When the error rate is high, it takes considerably more steps to reach the final accuracy level. This suggests another compute-error rate tradeoff that we do not explore in this paper, which is the steps it takes to a certain accuracy and the error rate of the verifier. Also, the rollout influences the stability of the accuracy curve, especially when the error rate is high. This shows another reason, other than the final accuracy and the speed of learning, why it is better to have a large number of rollouts when the verifier is noisy.

\end{document}